\documentclass{article}
\usepackage{ijcai21}

\usepackage{times}

\usepackage{soul}
\usepackage{url}
\usepackage[hidelinks]{hyperref}
\usepackage[utf8]{inputenc}
\usepackage[small]{caption}
\usepackage{graphicx}
\usepackage{amsmath}
\usepackage{amssymb}
\usepackage{booktabs}
\newtheorem{definition}{Definition}
\newtheorem{problem}{Problem}
\usepackage{color}
\usepackage{xcolor}

\usepackage{enumitem}
\setlist{leftmargin=*}
\usepackage{xspace}
\newcommand{\method}{\texttt{MICRON}\xspace }

\title{{\it Change Matters:} Medication Change Prediction with Recurrent Residual Networks}

\author{
Chaoqi Yang$^1$\and
Cao Xiao$^2$\and
Lucas Glass$^{2}$\And
Jimeng Sun$^{1*}$\\
\affiliations
$^1$Department of Computer Science, University of Illinois Urbana Champaign\\
$^2$Analytics Center of Excellence, IQVIA \\
\emails
chaoqiy2@illinois.edu,
cao.xiao@iqvia.com,
Lucas.Glass@iqvia.com,
jimeng@illinois.edu
}

\begin{document}

\maketitle
\begin{abstract}

Deep learning is revolutionizing predictive healthcare, including recommending medications to patients with complex health conditions.
Existing approaches focus on predicting all medications for the current visit, which often overlaps with medications from previous visits.
A more clinically relevant task is to identify {\it medication changes}.

In this paper, we propose  a new
recurrent residual networks, named \method, for medication change prediction. \method takes the changes in patient health records as input and learns to update a hidden medication vector and the medication set recurrently with a reconstruction design. The medication vector is like the memory cell that encodes longitudinal information of medications. Unlike traditional methods that require the entire patient history for prediction, 
\method has a residual-based inference that allows for sequential updating based only on new patient features (e.g., new diagnoses in the recent visit), which is efficient. 

We evaluated \method on real inpatient and outpatient datasets.  \method achieves $3.5$\% and $7.8$\% 
relative improvements over the best baseline in F1 score, respectively. \method also requires fewer parameters, which significantly reduces the training time to $38.3$s per epoch with $1.5\times$ speed-up.
\end{abstract}

\section{Introduction}

Recently years, deep learning has demonstrated initial success in potentially assisting clinical decision-making ~ \cite{almirall2012designing,choi2017using,xiao2018opportunities,mao2019medgcn}. Among others, the medication recommendation task has drawn lots of research interest~\cite{wang2017safe,wang2019order,shang2019pre,shang2019gamenet,zhang2017leap,killian2019learning,wang2018supervised}.
The common strategy of medication recommendation learns representations for medical
entities (e.g., diagnoses, medications) from
electronic health records, and use the learned representations to predict
medications that fit the patient’s health condition while avoiding adverse drug interactions.

% is to automatically suggest appropriate medications for treating the medical conditions of a patient while preventing the presence of adverse drug-drug interaction (DDI).

\begin{figure}[tbp!]
	\centering
	\includegraphics[width=3.3in]{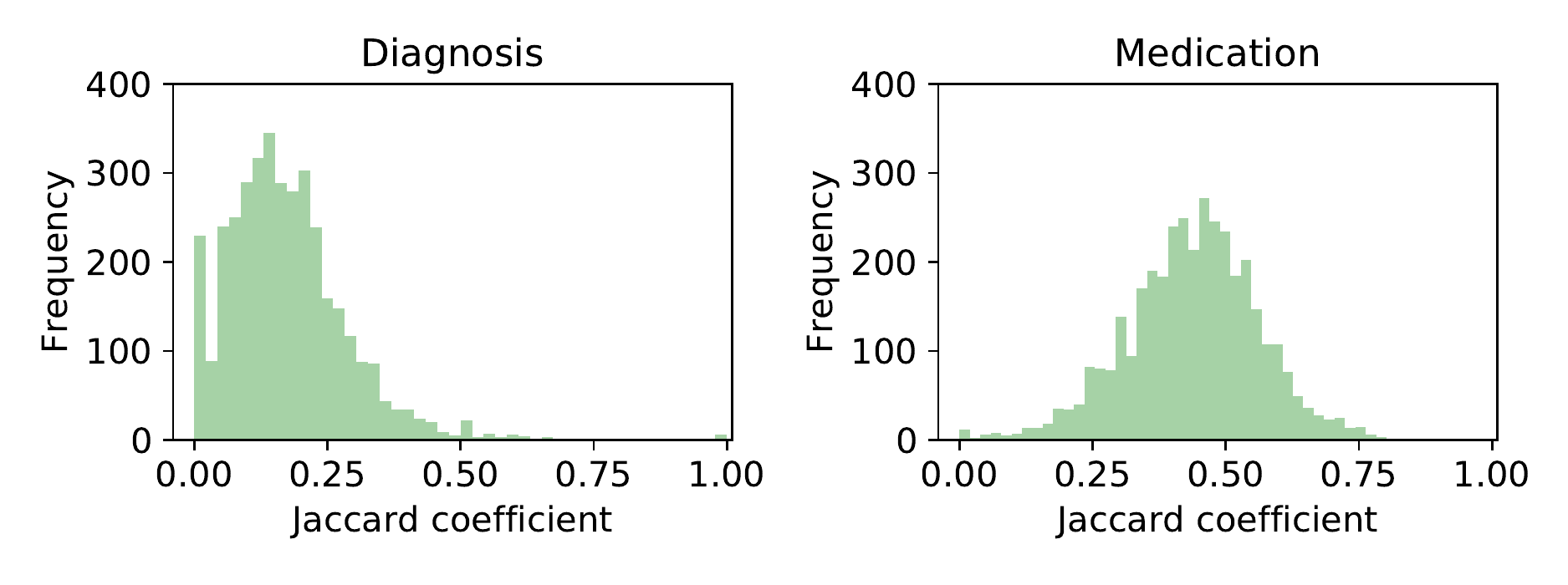}	\vspace{-3mm}
	\caption{Histogram of Jaccard Coefficients between consecutive visits. We observe relatively weaker overlaps in diagnoses but much stronger overlaps in medications over consecutive visits. It implies that the medication change is potentially more meaningful to predict.  %There are totally $4233$ samples and 50 bins. Each single sample represents one patient: for a patient's multiple visits, we first calculate a list of Jaccard coefficient between two consecutive visits, and then take the average.
	}
	\vspace{-3mm}
	\label{fig:jaccard_distribution}
\end{figure}

Many existing works focus on recommending the full set of medications in a visit~\cite{zhang2017leap,shang2019pre,shang2019gamenet,xiao2018opportunities}, which can be quite redundant from previous visits. Because the medications often remain stable over time with large overlaps between consecutive visits. For example, we investigate the Jaccard coefficient over consecutive visits on {\em MIMIC-III} data \cite{johnson2016mimic} (Fig.~\ref{fig:jaccard_distribution}). Among patients with multiple visits, although most of them are diagnosed with different conditions during consecutive visits (mean Jaccard below 0.2), the sets of medications remain stable (mean Jaccard around 0.5).%, explaining that the same underlying diseases could cause different symptoms. Still, the underlying diseases usually do not go away soon.

However, such medication patterns were rarely explored and leveraged to augment medication recommendation tasks. Challenges mainly arise from (1) how to accurately characterize the changes of patient health condition for each time step, and  (2) how to correctly identify the medication changes based on health changes.

To fill in the gap, we propose a new recurrent residual learning approach, named MedicatIon Change pRedictiON (\method) to predict medication changes and simultaneously model longitudinal medical history.  \method is enabled by the following technical contributions.

\begin{itemize}
\item {\bf Efficient representation of changing health conditions}. \method uses a residual health representation to sequentially update changes in patient health conditions, which provides more efficient model inference than RNN-based alternatives.
\item {\bf Explicit change set prediction:} \method decomposes the medication change prediction task into predicting two sets: (1) the {\em removal set} that removes previous medicines that are no longer needed; and (2) the {\em addition set} that brings in new medicines for newly developed diseases. Two sets are modeled by a recurrently updated medication vector and addition and removal thresholds selected at the high-confidence region from validation ROC-curve and thus could provide reliable inclusion-exclusion criterion.
\end{itemize}
We evaluated \method against state-of-the-art models on two real-world patient datasets: one inpatient {\em MIMIC-III} data and one outpatient dataset. \method outperforms the best baseline GAMENet \cite{shang2019gamenet} with $3.5$\% and $7.8$\% relative improvement in F1 measure, respectively. In addition, \method achieves $1.5\times$ speed-up in training and inference compared with GAMENet.

\vspace{-3mm}
\section{Related Works}
% Existing medication recommendation models mainly include the following three types.

\noindent
\textbf{Rule-based models} \cite{almirall2012designing,chen2016physician} typically rely on human-designed clinical guidelines, which require huge efforts from clinicians. For example, \cite{lakkaraju2017learning} optimizes a sequence of if-then-else rules, which maps the patient status into the prescription decision. 

\smallskip
\noindent
\textbf{Instance-based methods} extract patient features only from current visits. \cite{zhang2017leap} formulated the medication recommendation as a multi-instance multi-label (MIML) task and proposed a content-attention mechanism-based sequence-to-sequence model. \cite{wang2017safe} jointly embedded diseases, medicines, patients, and their corresponding relations into a shared space by the knowledge graph, which requires multiple external data sources.

\smallskip
\noindent
\textbf{Longitudinal approach} \cite{wang2018supervised,wang2019order,xiao2018opportunities,bhoi2020premier} is a popular approach that captures the sequential dependency in patient treatment history. \cite{choi2016retain,bajor2016predicting} modeled the longitudinal patient history by RNNs for various clinical predictive tasks. \cite{shang2019gamenet} and \cite{le2018dual} adopted memory-based networks with RNNs to handle the dependency among longitudinal medical codes.

% \noindent 
Compared with existing works, \method is based on a new perspective that focuses on predicting the changes. This is  more realistic since clinicians usually update 
patient prescriptions by only a small proportion for a patient's new visit.

\vspace{-0.5mm}
\section{Method}
% \subsection{Background} Drug replacement opens new opportunities for drug recommendation task: in the latter case, the models are required to predict a whole drug set, whereas in the former case, the sequential correlations between drug sets could be utilized and the task is barely of updating not predicting the whole drug sets between consecutive visits. The underlining assumption is that a drug set could be decomposed into two parts: (i) out-of-date (old) drugs; (ii) chronic (persistent) drugs. During two consecutive visits, we assume that the chronic parts will remain the same, since those related chronic diseases will not been cured suddenly. The old drugs will be discarded in the next visit, while we should introduce new drugs for current potentially new disease conditions. In sum, during two consecutive visits, models need to detect and discard some old drugs from the previous drug set and include new drugs to form the updated drug set.

\subsection{Problem Formulation}
\label{sec:notations}

\begin{definition}[Patient EHR Records]
Patient EHR records are usually represented by an ordered sequence of tuples. For a patient $j$, we denote his/her clinical documentaries as $\mathbf{X}_{j}=[\mathbf{x}_{j}^{(1)},\mathbf{x}_{j}^{(2)},\mathbf{x}^{(3)}_j,\dots]$, where the $t_{th}$ entry, $\mathbf{x}^{(t)}_j$, records the information of the $t_{th}$ visit, such as diagnoses, procedures and prescription information. In the paper, $\mathbf{x}^{(t)}_j=[\mathbf{d}^{(t)}_j,\mathbf{
p}^{(t)}_j,\mathcal{M}^{(t)}_j]$, where $\mathbf{d}^{(t)}_j\in\{0,1\}^{|\mathcal{D}|}$ and $\mathbf{p}^{(t)}_j\in\{0,1\}^{|\mathcal{P}|}$ are multi-hot diagnoses and procedure vectors, while $\mathcal{D}$ and $\mathcal{P}$ are the overall diagnosis and procedure sets.
%We refer diagnoses and procedures as clinical measurements, which are regarded as health features. 
$\mathcal{M}^{(t)}_j\subset\mathcal{M}$ is the $t_{th}$ medication set and $\mathcal{M}$ is a set for all possible medicines. We denote the visit-wise medication addition (new) and removal (old) sets as $\mathcal{N}^{(t)}_{target},\mathcal{O}^{(t)}_{target}\subset \mathcal{M}$, separately, which naturally follows the equality, ${\mathcal{M}}^{(t)} = ({\mathcal{M}}^{(t-1)} \cup \mathcal{N}^{(t)}_{target})\setminus \mathcal{O}^{(t)}_{target}$. 
\end{definition}
\vspace{-0.5mm}

% Adverse drug-drug interactions (DDIs) are a main challenge for drug co-prescription. In this paper, we refer the DDI relations to a survey \cite{tatonetti2012data}  and represent it by matrix $\mathbf{A}\in\{0, 1\}^{|\mathcal{M}|\times |\mathcal{M}|}$, where $\mathbf{A}_{ij}$ implies that medicine $i$ and $j$ could interact.

% Our formulated {\em medication change prediction} is originated form medication recommendation, which is usually formulated as: for one patient, given his/her longitudinal heath information, the model is required to generates visit-level medication recommendations. 
\begin{problem}[Medication Change Prediction]
Medication change prediction aims at determining the medication addition set $\mathcal{N}^{(t)}$ and the removal set $\mathcal{O}^{(t)}$ at visit $t$, given last prescription, $\tilde{\mathcal{M}}^{(t-1)}$ and patient health history $[\mathbf{d}^{(1)},\dots,\mathbf{d}^{(t)}]$ and $[\mathbf{p}^{(1)},\dots,\mathbf{p}^{(t)}]$.
%the model will predict medication changes: an addition set $\mathcal{N}^{(t)}$ and a removal set $\mathcal{O}^{(t)}$. Thus, the recommended medications of the $t_{th}$ visit are given by $\tilde{\mathcal{M}}^{(t)} = (\tilde{\mathcal{M}}^{(t-1)} \cup \mathcal{N}^{(t)})\setminus \mathcal{O}^{(t)}$. 
%where $\mathcal{A}\cup \mathcal{B}$ is the set union operation and $\mathcal{A}\setminus \mathcal{B}$ is set subtraction. 
The model aims to minimize the gap between current estimation $\tilde{\mathcal{M}}^{(t)}= (\tilde{\mathcal{M}}^{(t-1)} \cup \mathcal{N}^{(t)})\setminus \mathcal{O}^{(t)}$ and real prescriptions ${\mathcal{M}}^{(t)}$, and also control the incidence of DDIs
% \cite{tatonetti2012data} 
as denoted by $\mathbf{A}\in\{0, 1\}^{|\mathcal{M}|\times |\mathcal{M}|}$, where $\mathbf{A}_{ij}=1$ implies that medicine $i$ and $j$ could interact. 
\end{problem}

\subsection{\method Method}

\begin{figure*}[!ht]
	\centering
	\includegraphics[width=6.3in]{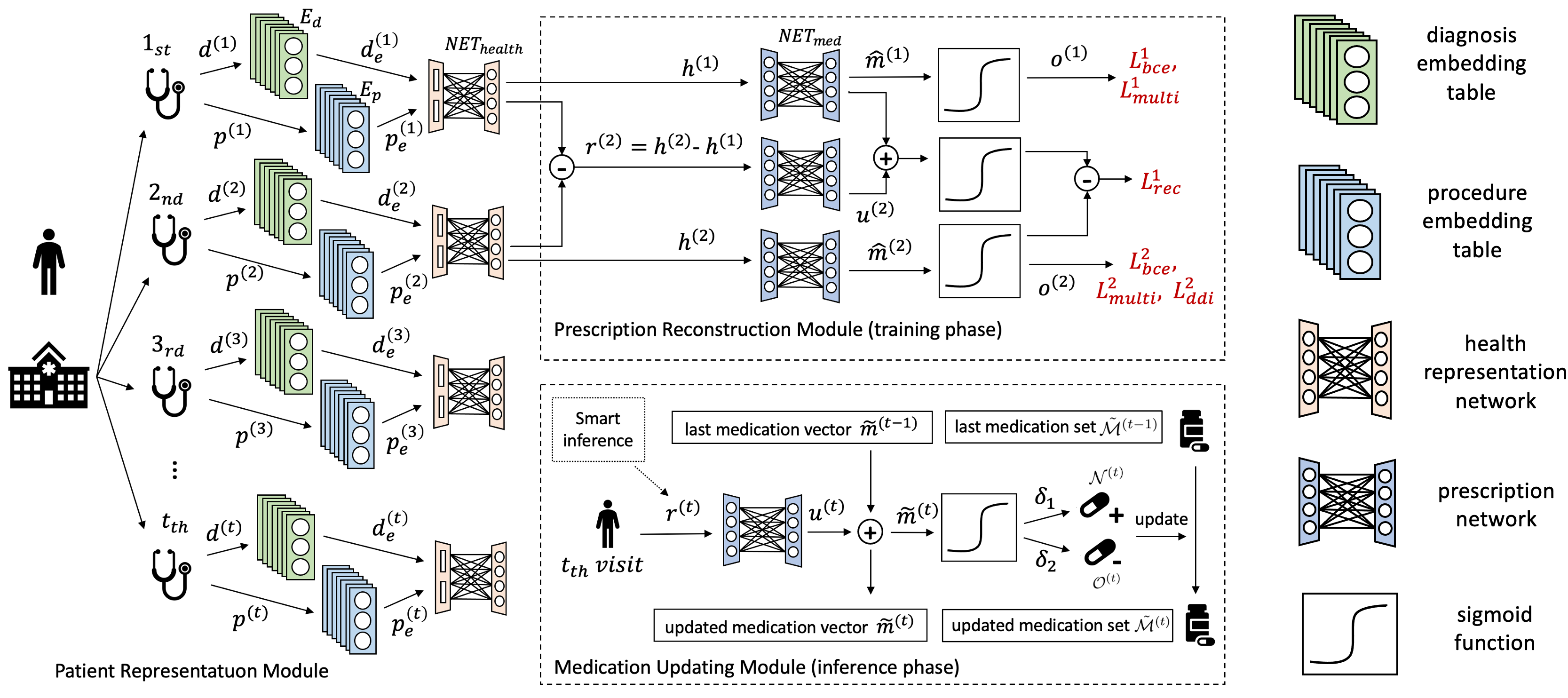}
	\vspace{-1mm}
	\caption{\method Framework. To represent a patient health condition, the model first feeds on diagnosis and procedure information and then generates a compact patient health representation by {\em health representation network}, an affine function. During training, the model uses a feed-forward network for {\em prescription network} and learns the residual representation under a novel reconstruction loss. In the inference, our model inputs the heath update to the same {\em prescription network} and then generates addition/removal sets to update the current prescription.}
	\label{fig:framework}
	\vspace{-1mm}
\end{figure*}

{\bf Overview}
%Our model is developed because the consecutive medication sets for the same patient often have strong overlaps.
As shown in Fig.~\ref{fig:framework}, \method has three modules: (1) a {\em patient representation module} that embeds  diagnosis and procedure codes into latent health representation; (2) a {\em prescription reconstruction module (training phase)}, where \method trains on consecutive pairs of visits and learns residual medication representations under a new reconstruction design;
(3) a {\em medication updating module (inference phase)} for model inference, where \method
initializes with previous medication information. For each subsequent visit, \method only requires an update of patient health status and then will predict the changes in the existing medications.

The key difference between \method and existing medication recommendation models \cite{shang2019gamenet,choi2016retain} is that while these models learn global sequential patterns using RNNs, \method learns sequential information locally (by every two consecutive visits) and propagates them visit-by-visit to preserve the longitudinal patient information.
% During a subsequent visit, \method starts from initial medication information, a real-valued {\em medication vector} and a {\em medication set}, which are provided by the last visit.
% We then utilize only the residual health representation to update the medication information, which is more efficient.

\subsection{Patient Representation}
Patient representation aims to learn a compact and indicative vector to represent a patient's status. In a clinical visit, doctors will recommend medications based on diagnosis and procedure information. Our module also feeds on these two features. Since \method is proposed for generic patients, we leave the subscript notation in the following.
% Note that, though we describe $d_i,p_i$ as sets in previous sections, in real practice, they should been transformed into multi-hot vectors before training. To make it clear, we have $d_i\in\{0,1\}^{|\mathcal{D}|}$ and  $p_i\in\{0,1\}^{|\mathcal{P}|}$.

\smallskip
\noindent\textbf{Diagnosis and Procedure Encoders.} For the $t_{th}$ visit, the input features, $\mathbf{d}^{(t)} \in\mathbb{R}^{|\mathcal{D}|}$ and $\mathbf{p}^{(t)} \in\mathbb{R}^{|\mathcal{P}|}$, can be extracted from clinical documentary. $\mathbf{d}^{(t)}$ is the multi-hot diagnosis vector, while $\mathbf{p}^{(t)}$ is the procedure vector. Following the similar strategy in \cite{zhang2017leap,shang2019gamenet}, we transform these two vectors into the embedding space using mapping matrices $\mathbf{E}_d\in\mathbb{R}^{s\times |\mathcal{D}|}$ and $\mathbf{E}_p\in\mathbb{R}^{s\times |\mathcal{P}|}$ ($s$ is the size of embedding space),
\begin{align}
\mathbf{d}^{(t)}_{e} =  \mathbf{E}_d \mathbf{d}^{(t)}~~~~\mbox{and}~~~~\mathbf{p}^{(t)}_{e} = \mathbf{E}_p \mathbf{p}^{(t)}.
\end{align}
%where each row of $\mathbf{E}_d$ ($\mathbf{E}_p$) is an embedding vector for an individual diagnosis (procedure). 
During training, these two tables are shared among all visits and patients.
The results, $\mathbf{d}^{(t)}_{e}$ and $\mathbf{p}^{(t)}_{e}$, are of the same dimension $\mathbb{R}^s$.

\smallskip
\noindent\textbf{Patient Hidden Representation.} To achieve one compact health representation, $\mathbf{d}^{(t)}_{e}$ and  $\mathbf{p}^{(t)}_{e}$ are further concatenated and parametrized by a {\em health representation network}, $\mbox{NET}_{health}$,
\begin{equation} \label{eq:patient_rep}
    \mathbf{h}^{(t)} = \mbox{NET}_{health}~\left(\left[\mathbf{d}^{(t)}_{e}~\|~\mathbf{p}^{(t)}_{e}\right]\right),
\end{equation}
which outputs an integrated health representation $\mathbf{h}^{(t)}\in\mathbb{R}^s$. In this paper, we use affine function (one layer neural network without activation) for $\mbox{NET}_{health}$.

Unlike previous works \cite{shang2019gamenet,zhang2017leap}, this paper does not use recurrent neural networks (RNN) to capture a patient's health history. 
%Our model is new in that we implicitly model the longitudinal patient health by a reconstruction design during training. 
Starting from $\mathbf{h}^{(t)}$, the model architecture differs in training and inference. Next, we elaborate on these two phases.

\subsection{Training: Prescription Reconstruction Module}
% \cx{highlight the novel design and their goal in the titles}
 Our \method trains on two consecutive visits, e.g., the $(t-1)_{th}$ and the $t_{th}$ visits. Given the health representations, i.e., $\mathbf{h}^{(t-1)}$ and $\mathbf{h}^{(t)}$, a straightforward way for recommending medications is to learn a mapping, i.e., a {\em prescription network}, $\mbox{NET}_{med}: \mathbb{R}^s \mapsto \mathbb{R}^{|\mathcal{M}|}$, from hidden embedding space to medication space for two visits, separately. 
\begin{align}
    \hat{\mathbf{m}}^{(t-1)} &= \mbox{NET}_{med}~(\mathbf{h}^{(t-1)}), \label{eq:complete1}\\
    \hat{\mathbf{m}}^{(t)} &= \mbox{NET}_{med}~(\mathbf{h}^{(t)}). \label{eq:complete2}
 \end{align}
where $\hat{\mathbf{m}}^{(t-1)},\hat{\mathbf{m}}^{(t)}\in\mathbb{R}^{|\mathcal{M}|}$ are medication representations and each entry quantifies a real value for the corresponding medicine. In the paper, $\mbox{NET}_{med}$ is implemented as a fully connected neural network.
To obtain the actual recommendations, a trivial way \cite{shang2019pre,shang2019gamenet} is to apply a {\em medication output layer}, which consists of a {\em Sigmoid} function $\sigma(\cdot)$, followed by a pre-defined threshold $\delta$, picking up medicines with larger activation value. However, in this paper, we hope to utilize and emphasize the dependency usage between $\mathbf{h}^{(t-1)}$ and $\mathbf{h}^{(t)}$ in 
the model.
% More concretely, the {\em medication output layer} consists of a {\em Sigmoid} function $\sigma(\cdot)$, followed by a pre-defined threshold $\delta$, picking up medicines with larger activation value.\js{what is the architecture of NET2}\\

\smallskip
\noindent\textbf{Residual Medication Representation.}
Formally, the difference between $\mathbf{h}^{(t-1)}$ and $\mathbf{h}^{(t)}$, i.e., $\mathbf{r}^{(t)} = \mathbf{h}^{(t)}-\mathbf{h}^{(t-1)}$, is denoted as \textit{residual health representation}, which encodes the changes in clinical health measurements, indicating an update in patient's health condition. Naturally, the health update $\mathbf{r}^{(t)}$ will cause an update in the resulting medication representation $\mathbf{u}^{(t)}$. Our motivation is that if $\mbox{NET}_{med}$ can map a complete health representation (e.g., $\mathbf{h}^{(t)}$) into a complete medication representation (e.g., $\hat{\mathbf{m}}^{(t)}$), then a residual health representation should also be mapped into an update in the same representation space through $\mbox{NET}_{med}$. In other words, $\mathbf{r}^{(t)}$ and $\mathbf{u}^{(t)}$ shall also follow the same mapping function, $\mbox{NET}_{med}$,
\begin{equation}
    \mathbf{u}^{(t)} = \mbox{NET}_{med}~(\mathbf{r}^{(t)}). \label{eq:residual}
\end{equation}
To learn Eqn.~\eqref{eq:complete1} and \eqref{eq:complete2}, we could use the medication combinations in the dataset as supervision, however, it is hard to formulate direct supervision for Eqn.~\eqref{eq:residual}. A simple idea is to model the addition and the removal medication sets separately (as we show in the experiment that separate modeling DualNN does not work well). Therefore, we consider reconstructing ${\mathbf{u}}^{(t)}$ from $\hat{\mathbf{m}}^{(t-1)}$ and $\hat{\mathbf{m}}^{(t)}$ by both unsupervised and supervised regularization.

% Task1 in this scenario is to learn an effective residual representation ${\mathbf{u}}^{(t)}$, without supervision from the dataset, and Task2 is to teach $\hat{\mathbf{m}}^{(t-1)}$ and $\hat{\mathbf{m}}^{(t)}$ for low DDI outcomes by real medication combinations.

\smallskip
\noindent\textbf{Unsupervised Residual Reconstruction.} 
% \js{Why this is main and the other is auxiliary? We need a sentence to justify the importance of this task over the other one. If they are of the same importance, we should just call them task 1 and 2 }
To model the medication changes, we design a reconstruction loss.
For Eqn.~\eqref{eq:complete1}, \eqref{eq:complete2} and  \eqref{eq:residual}, the inputs follow a residual relation: $\mathbf{h}^{(t-1)} + \mathbf{r}^{(t)} = \mathbf{h}^{(t)}$. Naturally, we also impose a similar relation in the {\em medication output layer} by introducing an {\em unsupervised reconstruction loss} ($\sigma(\cdot)$ is a {\em Sigmoid} function), 
\begin{equation}
    L^{(t)}_{rec} = \|\sigma(\hat{\mathbf{m}}^{(t-1)}+{\mathbf{u}}^{(t)}) - \sigma(\hat{\mathbf{m}}^{(t)})\|_2,
\end{equation}
which is calculated with $L_2$ norm. This reconstruction loss enforces the reconstructed recommendations from $\hat{\mathbf{m}}^{(t-1)}$ and the residual ${\mathbf{u}}^{(t)}$ to be close to the recommendations given by $\hat{\mathbf{m}}^{(t)}$. We show in the experiment that $L^{(t)}_{rec}$ is essential for learning the residual.

\smallskip
\noindent\textbf{Supervised Multi-label Classification.} To jointly modeling a low DDI output, we introduce three differentiable loss functions to improve $\hat{\mathbf{m}}^{(t-1)}$ and $\hat{\mathbf{m}}^{(t)}$, so as to achieve a better reconstruction ${\mathbf{u}}^{(t)}$.
% previous work \cite{choi2016retain,shang2019gamenet} has formulated the problem as a multi-label classification problem, where the prediction for each drug is an independent binary classification.

\begin{itemize}
    \item {\em Drug-Drug Interaction Loss.} Since adverse drug-drug interaction (DDI) is a leading cause of  morbidity and mortality in clinical treatments \cite{percha2013informatics}, we penalize the presence of DDIs in the output medication representation, $\hat{\mathbf{m}}^{(t)}$. First, we transform it by {\em Sigmoid} function, $\hat{\mathbf{o}}^{(t)} = \sigma(\hat{\mathbf{m}}^{(t)})$, and then design the DDI loss as,
\begin{equation}
    L^{(t)}_{ddi} = \sum_{i=1}\sum_{j=1} \mathbf{A}_{ij} \cdot \hat{\mathbf{o}}^{(t)}_i \cdot \hat{\mathbf{o}}^{(t)}_j,
\end{equation}
where $\mathbf{A}$ is the binary DDI matrix, extracted externally \cite{tatonetti2012data} and $\mathbf{A}_{ij}$ indicates that medicine $i$ and $j$ have interaction or not. The term $\mathbf{A}_{ij} \cdot \hat{\mathbf{o}}^{(t)}_i \cdot \hat{\mathbf{o}}^{(t)}_j$ is the a scalar product, which is the interaction penalty for medicine $i$ and $j$, and $\hat{\mathbf{o}}^{(t)}_i$ is the $i$-th element of the vector. Since we care about the DDI rate in the reconstructed representation, this loss only applies to the current visit $t$.
\item {\em Binary Cross-entropy Loss.} In addition, we also extract real medication set as supervision. Assume a multi-hot vector ${\mathbf{m}}^{(t)}\in\{0,1\}^{|\mathcal{M}|}$ is the vectorization of the target medication set $\mathcal{M}^{(t)}$. We adopt {binary cross entropy (BCE) loss},
% and $\mathcal{M}_*$ is the overall drug set
\begin{equation} \footnotesize
    L^{(t)}_{bce} = -\sum_{i=1} {\mathbf{m}}^{(t)}_i log (\hat{\mathbf{o}}^{(t)}_i)+(1-{\mathbf{m}}^{(t)}_i)log (1-\hat{\mathbf{o}}^{(t)}_i),
\end{equation}
where subscript $i$ indicates each element of the vectors. For this loss function, we compute on both $\hat{\mathbf{m}}^{(t-1)}$ and $\hat{\mathbf{m}}^{(t)}$.
\item {\em Multi-Label Margin Loss.} Then, we employ margin-based loss to enlarge the gap between the recommended medications and the unselected ones. Since  $\hat{\mathbf{o}}^{(t)}_i\in(0,1)$, the margin is set 1 in our paper.
\begin{equation}\vspace{-1mm} \footnotesize
L^{(t)}_{multi} = \sum_{i,j:~\mathbf{m}^{(t)}_i=1,\mathbf{m}^{(t)}_j\neq 1}
\frac{\mbox{max}(0,1-(\hat{\mathbf{o}}^{(t)}_i-\hat{\mathbf{o}}^{(t)}_j))}{|\mathcal{M}|}. 
\end{equation}
We also consider to penalize both of the visits, i.e., calculating $L^{(t-1)}_{multi}$ and $L^{(t)}_{multi}$ using this loss.
\end{itemize}
These three losses use external supervision to optimize the {\em prescription network}, so that during inference, our \method would predict medication changes more accurately.

\smallskip
\noindent\textbf{Overall Loss Function.}
In the training process, we hope to find optimal values for embedding tables, $\mathbf{E}_d$ and $\mathbf{E}_p$, parameter matrices in $\mbox{NET}_{health}$ and $\mbox{NET}_{med}$. The loss functions are combined by weighted sum,
\begin{align}
    L_{total} &= \lambda_1 L^{(t)}_{rec}
    + \lambda_2 L^{(t)}_{ddi} + \lambda_3\left(\gamma L^{(t)}_{bce} + (1-\gamma)L^{(t-1)}_{bce}\right) \notag\\
    & 
    + \lambda_4\left(\gamma L^{(t)}_{multi} + (1-\gamma)L^{(t-1)}_{multi}\right), \label{eq:loss}
\end{align}
where $\lambda_i,~i=1,2,3,4$, are different weights for four types of loss functions, and $\gamma$ is introduced to balance two consecutive visits. During the training, one batch contains all visits of one patient, and the loss is back-propagated after each batch. In the paper, we treat the weights as hyperparameters. In Appendix, we also prototype a momentum-based method to select the weights automatically.

% We select $\gamma$ before the training. For $\lambda_i,~i=1,2,3,4$, we design momentum-based weights, which is adjusted adaptively before loss back-propagation.

\vspace{1mm}

% These five factors are tuned in validation data.
% \js{all these hyperparameters beg the questions how to effectively choose them. We need to provide some heuristics here. Otherwise this will be a weakness of our method as one might expect the performance may be sensitive to the choices of those weight parameters in the loss}
\subsection{Inference: Medication Updating Module}
% During the inference phase, we follow the concept of drug replacement. Replacement is more realistic than per-visit drug recommendations.

To predict medication changes, it is essential to maintain a {\em medication combination}. We hope that for the subsequent visit, it would be enough to derive an update in the combination based on new diagnosis or procedure information. However, like Risperdal for treating schizophrenia and bipolar disorder, some medicines will not be prescribed based on only one or two visits, and it might need long-term clinical observation. We therefore also maintain a {\em medication vector}, where each element quantifies the cumulative effect of a medicine. After clinical visits, each element in the vector will increase or decrease based on the updates of patient health status. Essentially, the {\em medication vector} is like the memory cell in RNNs, which is refreshed visit-by-visit. Once it is above or below certain thresholds, the medicine will be added or removed from the current sets. More concretely, the medicine change prediction follows three steps.

\smallskip
\noindent\textbf{Step 1: Medication Vector Update.} Specifically, for the $t_{th}$ visit of a patient, the medication changes start from a medication vector, $\tilde{\mathbf{m}}^{(t-1)}\in\mathbb{R}^{|\mathcal{M}|}$, and a medication set, $\tilde{\mathcal{M}}^{(t-1)}\subset\mathcal{M}$. The model first updates the vector based on a residual health representation, $\mathbf{r}^{(t)}$,
\begin{align}
    \tilde{\mathbf{m}}^{(t)} &= \tilde{\mathbf{m}}^{(t-1)} + {\mathbf{u}}^{(t)} \notag\\ 
    & = \tilde{\mathbf{m}}^{(t-1)} + \mbox{NET}_{med}~(\mathbf{r}^{(t)}),
\end{align}
where $\mathbf{r}^{(t)}$ is calculated by $\mathbf{h}^{(t)}-\mathbf{h}^{(t-1)}$ (defined in Eqn.~\eqref{eq:patient_rep}, $\mbox{NET}_{health}$), which is implemented as an affine function. We use an efficient {\em smart inference} module to calculate $\mathbf{r}^{(t)}$, in case only the updates in medical codes (e.g., diagnosis and procedure) are accessible. We specify it in Appendix.

\smallskip
\noindent\textbf{Step 2: Addition and Removal.} Then, based on the updated medication vector, $\tilde{\mathbf{m}}^{(t)}\in\mathbb{R}^{|\mathcal{M}|}$, we identify which medicines are ready to add or remove. We design two thresholds ($\delta_1$ is for the addition set, while $\delta_2$ is for the removal set, where $1\geq\delta_1\geq \delta_2\geq0$) to control the size of changes. Specifically, we first apply a {\em Sigmoid} function $\sigma(\cdot)$, and then the addition and removal sets are generated by applying the thresholds $\delta_1$ and $\delta_2$ element-wise,
\begin{align}
    \mathcal{N}^{(t)} &= \{i\mid \sigma(\tilde{\mathbf{m}}^{(t)}_i) \geq\delta_1\}, \\
    \mathcal{O}^{(t)} &= \{i\mid \sigma(\tilde{\mathbf{m}}^{(t)}_i) \leq \delta_2\},
\end{align}
where $\mathcal{N}^{(t)}$ ($\mathcal{O}^{(t)}$) is for addition (removal) set, and subscript $i$ enumerates the index of $\tilde{\mathbf{m}}^{(t)}$. Note that, $\mathcal{N}^{(t)}\cap \mathcal{O}^{(t)} = \varnothing$. For two thresholds, if $\delta_1=1$ and $\delta_2=0$, then $\mathcal{N}^{(t)} = \mathcal{O}^{(t)} = \varnothing$; in another case, $\delta_1=\delta_2$, then ``medication change prediction" becomes ``full medication prediction". 

The thresholds $\delta_1$ and $\delta_2$ are selected based on the receiver operating characteristic (ROC) of each medicine. Specifically, we load the pre-trained \method on the validation set. For each medicine, we collect cut-off thresholds of the ROC curve in descending order. $\delta_1$ is the average of $5$-percentile over all medications, while $\delta_2$ is based on $95$-percentile.
Essentially, $\delta_1$ will provide a low false negative (FN) rate, while $\delta_2$ ensures a low false positive (FP) rate.

% We use 95\% and 5\% quantiles of the  receiver operating characteristic (ROC) curve to select $\delta_1$ and $\delta_2$. It is specified in Appendix~B.\\

% However, $\mathcal{N}^{(t)}$ and $\hat{M}_{t-1}$ can have overlaps, while $o_t$ could also contain drugs that is not in  $\hat{M}_{t-1}$. These overlaps will not affect the final replacement results due to the set operations.

\smallskip
\noindent\textbf{Step 3: Medication Set Update.} Next, we apply the changes (addition and removal) in the existing medication set. The generation of a new combination is given by set operations,% (a \textit{Venn} illustration is shown in Fig.~\ref{fig:medication_change}),
\begin{equation}
    \tilde{\mathcal{M}}^{(t)} = (\tilde{\mathcal{M}}^{(t-1)} \cup {\mathcal{N}}^{(t)}) \setminus {\mathcal{O}}^{(t)},
\end{equation}
where we use set union and subtraction operation. ${\mathcal{N}}^{(t)}$ and $\tilde{\mathcal{M}}^{(t-1)}$ could have overlaps, while ${\mathcal{O}}^{(t)}$ could also contain medications that are not in $\tilde{\mathcal{M}}^{(t-1)}$. The overlaps will not affect the final recommendation results due to set operations. 

To sum up, the model begins with a  medication vector, $\tilde{\mathbf{m}}^{(t-1)}$, and a medication set, $\tilde{\mathcal{M}}^{(t-1)}$, which are provided by the previous visit. During the current visit, \method uses the update of patient status as input and walks through the above three steps to finish one round of medication change, as well as to update $\tilde{\mathbf{m}}^{(t)}$ and $\tilde{\mathcal{M}}^{(t)}$ for the next visit.

\vspace{-0.2mm}
\section{Experiments}
\vspace{-0.3mm}
We evaluate \method against several baselines in both inpatient and outpatient datasets. We focus on answering the following questions: % which provides the following results:
\begin{itemize}
    \item How does \method perform against the baselines in medication and change prediction?
    \item How does \method perform in model efficiency?
    \item How do different components in \method  contribute to accurate recommendations?
    % \item The residual part is important, it quantifies the relation between clinical measurements and individual drugs. We further provide an experience table for doctor's reference.
\end{itemize}

\vspace{-2mm}
\subsection{Experimental Setup}

\noindent\textbf{Dataset}.
We consider a benchmark inpatient dataset: {\em MIMIC-III} \cite{johnson2016mimic}, and a private outpatient dataset: {\em IQVIA PharMetrics Plus} (see processed statistics in Table~\ref{tb:dataset}). Details of dataset descriptions, preprocessing, hyperparameter selections can be found in Appendix.

\begin{table}[h!] \small \caption{Statistics of Datasets}
\vspace{-2.5mm}\centering
	\begin{tabular}{l|cc} \toprule \textbf{Items} & \textbf{MIMIC-III} & \textbf{IQVIA}\\ 
	\midrule  
	\# of visits & 14,960 & 30,794 \\ 
	\# of patients & 6,335 & 3,023 \\ 
	\# of diagnosis codes & 1,958 & 1,744 \\ 
	\# of procedure codes & 1,430 & 1,250 \\
	\# of medication codes & 131 & 155 \\
	\bottomrule \end{tabular}
	\label{tb:dataset}% \vspace{-3mm} 
	\end{table}

\noindent\textbf{Baselines}.
We consider the following baselines (SimNN and DualNN are designed by ourselves).
\begin{itemize}
    \item {\bf SimNN} use the same patient representation $\mathbf{h}^{(t)}$ as \method and then learns a simple 3-way classifier for each medicine (add, remove, and remain) with the cross-entropy loss. 
    \item {\bf  DualNN} also  starts from patient representation $\mathbf{h}^{(t)}$ and then diverges to two different neural networks. The first one is for addition and the second for removal. Each neural network classifier uses the binary cross-entropy loss.
	\item{\bf LEAP} \cite{zhang2017leap} is an instance-based approach that uses a sequence to sequence model with reinforcement aftermath fine-tuning. This method generates a list of medications based on the diagnoses in the same visit.
	\item{\bf RETAIN} \cite{choi2016retain} is a longitudinal predictive model, which designs a specialized attention model over RNN. It learns the temporal dependencies between clinical visits and makes medication recommendations.
	\item{\bf GAMENet} \cite{shang2019gamenet} is also a longitudinal model, which uses RNN, memory network, and graph neural network. It predicts medications using historical prescriptions as reference.
\end{itemize}

\noindent\textbf{Evaluation Strategy and Metrics}.
We use evaluation metrics such as  DDI rate, Jaccard Similarity, F1-score similar to evaluate the overall recommended medications, as other related works \cite{shang2019gamenet,zhang2017leap}.  Also, we design new error metrics to evaluate the accuracy of medication changes: {\em Err(add)} and {\em Err(remove)}.

\begin{figure}[t!]
	\centering
	\includegraphics[width=2.7in]{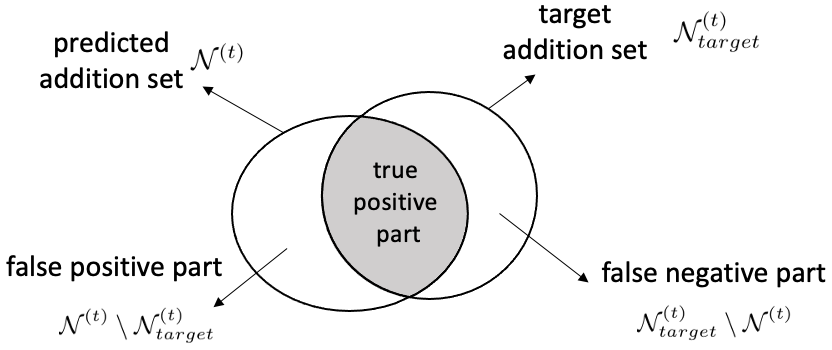}
	\caption{Illustration of Err(add) Metric at the $t$-th Visit}
	\label{fig:error_metric}
\end{figure}

{\em Err(add)} computes the sum of false positive part and false negative part (white region in Fig.~\ref{fig:error_metric}) between the predicted addition set $\mathcal{N}^{(t)}$ and the target addition set $\mathcal{N}_{target}^{(t)}$.  The target addition set is calculated by
\begin{equation}
     \mathcal{N}_{target}^{(t)} = \mathcal{M}^{(t)} \setminus \tilde{\mathcal{M}}^{(t-1)}, \notag
\end{equation}
where $\mathcal{M}^{(t)}$ is the target medication set in the $t_{th}$ visit, and $\tilde{\mathcal{M}}^{(t-1)}$ is the predicted medication set at the $(t-1)_{th}$ visit. For a particular patient $j$, this metric is calculated from the second visit, where the medication changes start,
{\footnotesize
\begin{align}
    \mbox{Err(add)}_j &= \frac{1}{V(j)}\sum_{t=2}^{V(j)} \left(|\mathcal{N}_{target}^{(t)}\setminus \mathcal{N}^{(t)}| + |\mathcal{N}^{(t)}\setminus \mathcal{N}_{target}^{(t)}|\right), \notag
\end{align}
}

\noindent where $\mathcal{A}\setminus\mathcal{B}$ is the set subtraction, and $V(j)$ means total number of visits of patient $j$. 
Finally, we average over all patients and get {\em Err(add)}. Similarly, we design {\em Err(remove)} to denote errors for removal. Also, we report model size and training/inference time. 
\vspace{-0.5mm}

 Since the evaluation is on medication change prediction, we assume the earliest visit appeared in the window as the ``first" visit of a patient $j$, where we could extract the initial medication set, $\tilde{\mathcal{M}}_j^{(1)}=\mathcal{M}_j^{(1)}$, and the initial medication vector, $\tilde{\mathbf{m}}_j^{(1)}=\hat{\mathbf{m}}_j^{(1)}$. For a fair comparison, the evaluation of all models starts from the ``second" visit. The definition of other metrics can be found in Appendix.

\subsection{Experimental Results}

\begin{table*}[!h]
    \caption{Performance Comparison (on {\em MIMIC-III / IQVIA})}
    \vspace{-2mm}
    \resizebox{\textwidth}{!}{\centering\begin{tabular}{c|ccccccc}
    \toprule
    \centering
      Metrics & DDI &Jaccard  & F1 & Err(add) & Err(remove) & Model Size & Train(epoch) /  Test\\
         \midrule
        %  Repeat & 0.0816 / 0.3062 & 0.4739 / 0.2872 & 0.6327 / 0.3676 & 7.301 / 1.626 & 6.875 / 0.783 & --- & --- \\
       SimNN & \underline{0.0837} / {0.0152} & 0.4658 / 0.1920 & 0.6235 / 0.2383 & {6.856} / 2.906 & 7.329 / 1.679 & 1.5MB (376,009 params) & 27.53s / 6.67s \\
       DualNN & 0.0925 / {0.0156} & 0.4880 / 0.1120 & 0.6447 / 0.1783 & \underline{6.261} / 3.406 & 7.987 / 2.579 & 1.3MB (325,702 params) & 27.90s / 5.76s \\
         RETAIN & 0.0932 / 0.0279 & 0.4796 / \underline{0.3320} & 0.6389 / \underline{0.4215} & {8.552} / \underline{2.353} & 6.338 / \underline{\bf 0.927} & 1.2MB (287,940 params) & 34.78s / 6.97s\\
         LEAP & {0.0880} / \underline{\bf 0.0134} & 0.4331 / 0.1871 & 0.5953 / 0.2742 & 9.105 / 3.663 & 5.939 / 1.286 & 1.7MB (433,286 params) & 199.94s / 26.45s \\
         GAMENet  & 0.0928 / 0.0166 & \underline{0.4980} / 0.2025 & \underline{0.6549} / 0.3016 & 8.810 / 3.016 & \underline{5.854} / 2.179  & 1.8MB (449,092 params) & 55.31s / 8.84s \\
         \midrule
         \method & {\bf 0.0695} / 0.0143  & {\bf 0.5234 / 0.3634} & {\bf 0.6778 / 0.4544} & {\bf 6.090} / {\bf 2.088} & {\bf 5.853} / 1.213 & { 1.1MB (275,395 params)} & { 38.83s / 4.41s}\\
         $\Delta$ Improve.  & $\downarrow17.0$\% / $\uparrow6.7$\% & $\uparrow5.1$\% / $\uparrow9.5$\% & $\uparrow3.5$\% / $\uparrow7.8$\% & $\downarrow2.7$\% / $\downarrow11.3$\% & $\downarrow0.02$\% / $\uparrow30.9$\% & --- & --- \\
         \bottomrule
    \end{tabular}}
    {\\~*The result of {\em MIMIC-III} and {\em DS2} are reported together and separated by ``/". The last two metrics, Model Size and Train(epoch)/Test, are based on {\em MIMIC-III} dataset. For the other five metrics, we select the best results in {\em bond font} and use {\em underscore} to select the best baseline. We also report the improvement $\Delta$ of our \method over the best baseline.}
    \label{tb:performance}

\end{table*}

\begin{table*}[!h] \centering
    \caption{Ablation Study for Different Model Components (on {\em MIMIC-III})}
    \vspace{-2mm}
    \resizebox{0.8\textwidth}{!}{\centering\begin{tabular}{l|ccccccc}
    \toprule
    \centering
      Metrics & DDI &Jaccard  & F1 & Err(add) & Err(remove) \\
         \midrule
        %  Repeat & 0.0816 / 0.3062 & 0.4739 / 0.2872 & 0.6327 / 0.3676 & 7.301 / 1.626 & 6.875 / 0.783 & --- & --- \\
       \method w/o $L_{rec}$ &  0.0618 $\pm$ 0.0002&  0.4449 $\pm$ 0.0138&  0.6050 $\pm$ 0.0014&  7.143 $\pm$ 0.3753&  6.224 $\pm$ 0.0598\\
\method w/o $\tilde{\mathbf{m}}^{(t)}$ &  0.0696 $\pm$ 0.0004&  0.4509 $\pm$ 0.0046&  0.6096 $\pm$ 0.0368&  8.496 $\pm$ 0.1401&  6.463 $\pm$ 0.1468\\
\method w/o $L_{multi}$ &  0.0780 $\pm$ 0.0015&  0.5020 $\pm$ 0.0072&  0.6590 $\pm$ 0.0288&  6.544 $\pm$ 0.2172&  5.509 $\pm$ 0.0732\\
\method w/o $L_{ddi}$ &  0.0931 $\pm$ 0.0005&  0.5248 $\pm$ 0.0006&  0.6793 $\pm$ 0.0081&  6.402 $\pm$ 0.3020&  5.897 $\pm$ 0.0397\\
\method w/o $\delta_1,\delta_2$ &  0.0628 $\pm$ 0.0018&  0.5074 $\pm$ 0.0016&  0.6635 $\pm$ 0.0026&  7.216 $\pm$ 0.1335&  5.084 $\pm$ 0.2706\\
\method & {0.0695} $\pm$ 0.0004  & {0.5234} $\pm$ 0.0008& {0.6778 } $\pm$ 0.0007& {6.090} $\pm$ 0.0189& {5.853} $\pm$ 0.0219\\
         \bottomrule
    \end{tabular}}
    \label{tb:ablation} \\
    For {\em DDI, Err(add), Err(remove)} metrics, the lower the better, while for {\em Jaccard} and {\em F1} metrics, the higher the better.
    \vspace{-2mm}
\end{table*}

We conduct experimental comparison based on five different random seeds and show the mean metric values in Table~\ref{tb:performance}.  \method outperforms all baselines in both inpatient and outpatient settings, especially for Jaccard and F1 metrics. LEAP gives a relatively good DDI measure on two datasets; however, its performance is weaker than other baselines in terms of accuracy. Although SimNN, DualNN, and RETAIN are implemented from very different perspectives, the former two are instance-based while the latter uses sequence modeling. They show neck-to-neck performance on {\em MIMIC-III}. For outpatient medication change prediction (on {\em IQVIA}), RETAIN shows strong performance while some recent state of the art baselines failed, such as GAMENet. We hypothesize that time spans between two visits can be much longer for outpatients, and thus the stored memory can be less trustworthy in GAMENet. By learning an effective residual representation, \method provides more accurate and safe medication recommendations for inpatient or outpatient settings. Also, \method requires much fewer parameters than the state-of-the-art approaches, which is more efficient.

Due to space limitation, the standard deviation results are reported in the Appendix. We also test the model's stability and do a T-hypothesis testing of \method on each metric. As a summary, most of the $p$-values are less than $0.001$ (mean $p$-value at 6.2e-5), except in two cases on {\em IQVIA}: the DDI rate compared to LEAP and the Err(remove) compared to RETAIN.

\subsection{Ablation Study on Model Components}

In this section, we verify the effectiveness of different components in \method. Specifically, we conduct ablation studies on {\em MIMIC-III} and test on the following variants: 
\begin{itemize}
    \item (i) \method { \em w/o $L_{rec}$}. We remove the unsupervised loss during training and solely trained on the supervised loss. 

    \item (ii) \method { \em w/o $\tilde{\mathbf{m}}^{(t)}$}. We do not maintain the medication vector, $\tilde{\mathbf{m}}^{(t)}$, and only utilizes the update feature information, $\mathbf{r}^{(t)}$, between two visits;

    \item (iii) \method { \em w/o $L_{multi}$}. We remove $L_{multi}$, and it will be less confident to use thresholds, $\delta_1$ and $\delta_2$; 

    \item (iv) \method { \em w/o $L_{ddi}$}. We remove DDI loss, and the model probably would provide high-DDI combinations;

    \item (v) \method { \em w/o $\delta_1,\delta_2$}. We set $\delta_1=\delta_2=\delta$, which implies medications with  score above or equal $\delta$ being added, and medications with score less than $\delta$ being removed.  This is a common strategy used in previous works: $\delta=0.5$ in \cite{shang2019gamenet} and $\delta=0.3$ in \cite{shang2019pre} (this model requires ontology information, so it is not included as baseline). We use $\delta=0.5$ for this model variant.

\end{itemize}
The comparison results with variances (after $\pm$) are shown in Table~\ref{tb:ablation}. 
Overall, all other variations perform better than variant (i) and (ii), highlighting that the reconstruction design and the initial medication vector are essential in the model. Without medication vector $\tilde{\mathbf{m}}^{(t)}$, the model cannot retain the longitudinal information, thus variant (ii) provides poor results. We also notice that without DDI loss, variant (iv) outputs a significantly higher DDI rate, and \method shows slightly better results than model variant (iii) without $L_{multi}$. By integrating all components, \method achieves a more balanced and stable performance in all metrics.

\section{Conclusion}

This paper tackles the medication change prediction problem and proposes a recurrent residual learning model, named \method, for predicting medication changes. We compare our model with state of the art approaches and show its effectiveness and efficiency on inpatient {\em MIMIC-III} dataset and a proprietary outpatient {\em IQVIA} dataset.

Like previous medication recommendation works,
% \cite{shang2019gamenet,zhang2017leap,shang2019pre},
this paper uses the existing prescriptions as a gold standard. The efficacy of the recommendation is evaluated by comparing it with the prescriptions given by the dataset, which might be a limitation. In the future, we will perform a clinical user study to evaluate our results further. Also, we would like to consider three practical extensions: (i) enhance the model by utilizing relations between medications, e.g., similarities; (ii) consider medicine dosages during recommendation; (iii) penalize different medications based on patient-specific conditions.

\bibliographystyle{named}
\bibliography{ijcai21}

\clearpage

\appendix

\section{Momentum-based Loss (MBL) Functions}

As stated in the main text, the weights for different loss, $\lambda_i,~i=1,2,3,4$, are treated as hyperparameters. To automate the procedure of hyperparameter search, we design momentum-based weights, which are adjusted adaptively during the training process. This module is primarily proposed for automatic selection of four hyperparameters, not for performance improvement.

{\em Momentum-based Weights.} The key advantage of using momentum is to combine current exploration with historical information. For each training epoch, we maintain a mean value for each type of loss, $\bar{L}_i^{(k)}$, which is calculated over the first $k$ training samples ($i=1,2,3,4$ stand for {\em rec, ddi, bce, multi} parts, separately). We use $L_i^{(k)}$ to denote the actual loss at the $k_{th}$ training point, and the momentum-based weights, $\lambda_i^{(k)}$, are calculated by the relative difference from the mean,
{\footnotesize
\begin{align}
    \vspace{-1mm}
    \mbox{diff}_i^{(k)} &= \frac{L_i^{(k)} - \bar{L}_i^{(k-1)}}{\bar{L}_i^{(k-1)}}, \\
    \vspace{-1mm}
    \lambda_i^{(k)} &= \gamma\cdot\frac{exp(\mbox{diff}_i^{(k)})}{\sum_{i}exp(\mbox{diff}_i^{(k)})} + (1-\gamma)\cdot\lambda_i^{(k-1)},
    \vspace{-1mm}
\end{align}
}%
where we use the same $\gamma$ (in Eqn.~\eqref{eq:loss}) to balance the current exploration and historical inertia. Meanwhile, the momentum values are also updated by,
\begin{equation}\footnotesize
    \bar{L}_i^{(k)} = \frac{L_i^{(k)} + (k-1)\cdot\bar{L}_i^{(k-1)}}{k}.
\end{equation}
The momentum weights are in the same spirit as the negative control. The momentum weight would enable our model to automatically identify and focus on the under-optimized parts in the loss as well as utilizing historical loss information. It empirically provides stable results. Note that in model implementation, we further employ a DDI threshold, $\eta\in(0,1)$. If the result DDI rate of the $k_{th}$ sample is below the threshold, we will directly set $\lambda_2^{(k)}=0$. We provide an ablation study on the momentum-based loss in Table~\ref{tb:MBL}.

\begin{table}[!htbp] \small
    \caption{Ablation Study on Momentum-based Loss (for MIMIC-III / DS2)}
    \vspace{-2mm}
    % \resizebox{90mm}{80mm}{
    \centering\begin{tabular}{c|c|c}
    \toprule
    Metrics & \method & \method with MBL \\
    \midrule
    DDI & 0.0695 / 0.0143  & 0.0690 / 0.0147 \\
    Jaccard & 0.5234 / 0.3634  & 0.5252 / 0.3479 \\
    F1 & 0.6778 / 0.4544 & 0.6807 / 0.4536 \\
    Err(add) & 6.090 / 2.088 & 6.255 / 2.115 \\
    Err(remove) & 5.853 / 1.213 & 5.623 / 1.157 \\
    \bottomrule
    \end{tabular}
    \label{tb:MBL}
    \vspace{-1mm}
\end{table}

\section{Smart Inference} This module will potentially improve the efficiency and interpretation of our model. Our model can reflect when each medicine is added or dropped and why (based on the appearance or disappearance of what diagnoses or procedures). Typically, during the inference phase,
$\mathbf{r}^{(t)}$ is calculated by
{\small \begin{align}
    \mathbf{r}^{(t)}&=\mathbf{h}^{(t)}-\mathbf{h}^{(t-1)} \notag\\
    &{ =\mbox{NET}_{health}\left(\left[\mathbf{d}^{(t)}_e~\|~\mathbf{p}^{(t)}_e\right]\right) - \mbox{NET}_{health}\left(\left[\mathbf{d}^{(t-1)}_e~\|~\mathbf{p}^{(t-1)}_e\right]\right)}\label{eq:affine} .
    \vspace{-2mm}
\end{align}}

\vspace{-2mm}
However, we can trace the cause back to the very beginning if $\mbox{NET}_{health}$ is \textit{affine}, 
\begin{equation}
    \mbox{NET}_{health}(\mathbf{x}-\mathbf{y}) = \mbox{NET}_{health}(\mathbf{x}) - \mbox{NET}_{health}(\mathbf{y}),~\forall \mathbf{x},\mathbf{y} \notag
\end{equation}
where Eqn.~\eqref{eq:affine} could be written as
{\small \begin{align}
    \mathbf{r}^{(t)}
    &={ \mbox{NET}_{health}\left(\left[\mathbf{d}^{(t)}_e-\mathbf{d}^{(t-1)}_e~\|~\mathbf{p}^{(t)}_e-\mathbf{p}^{(t-1)}_e\right]\right)} \notag \\ 
    &= { \mbox{NET}_{health}\left(\left[(\mathbf{d}^{(t)}-\mathbf{d}^{(t-1)})\cdot\mathbf{E}_d~\|~(\mathbf{p}^{(t)}-\mathbf{p}^{(t-1)})\cdot\mathbf{E}_p\right]\right)}\label{eq:fast}.
\end{align}}
% The input could be further derived from,
% {\small \begin{align}
%     \mathbf{d}^{(t)}_e-\mathbf{d}^{(t-1)}_e & =
%     \sum_{\{i: ~\mathbf{d}^{(t)}_i=1\}}\mathbf{E}^d_{i} - 
%     \sum_{\{i: ~\mathbf{d}^{(t-1)}_i=1\}}\mathbf{E}^d_{i} \notag\\
%     & = \sum_{\{i: ~\mathbf{d}^{(t)}_i=1, \mathbf{d}^{(t-1)}_i=0\}}\mathbf{E}^d_{i} - 
%     \sum_{\{i: ~\mathbf{d}^{(t-1)}_i=1, \mathbf{d}^{(t)}_i=0\}}\mathbf{E}^d_{i} \label{eq:diagnosis}\\
%     \mathbf{p}^{(t)}_e-\mathbf{p}^{(t-1)}_e & =
%     \sum_{i\in \mathcal{P}^{(t)}}\mathbf{E}^p_{i} - 
%     \sum_{i\in \mathcal{P}^{(t-1)}}\mathbf{E}^p_{i} \notag\\
%     & = \sum_{i\in \mathcal{P}^{(t)}\setminus \mathcal{P}^{(t-1)}}\mathbf{E}^p_{i} - 
%     \sum_{i\in \mathcal{P}^{(t-1)}\setminus \mathcal{P}^{(t)}}\mathbf{E}^p_{i} \label{eq:procedure}
% \end{align}}
In fact, this equation reveals the causality between the update in diagnosis and procedure measurements and the update in patient health representation. For example, a patient get extra X-ray scanning at this visit while other diagnosis and procedure measurements are the same as the last time. Then, the model input is barely the code of X-ray, and the input will hopefully stimulate some of the medicines while discourage others, based solely on the X-ray information.

Eqn.~\eqref{eq:fast} is practically efficient, since the clinical update $\mathbf{d}^{(t)}-\mathbf{d}^{(t-1)}$ (or $\mathbf{p}^{(t)}-\mathbf{p}^{(t-1)}$) is much sparser than original diagnosis documentary $\mathbf{d}^{(t-1)}$ and $\mathbf{d}^{(t)}$ (or procedure documentary $\mathbf{p}^{(t-1)}$ and $\mathbf{p}^{(t)}$). Also, it only requires one-time computation of $\mbox{NET}_{health}$ for the residual health representation. 

\section{Datasets and Hyperparameter Settings}
\begin{figure}[thbp!]
	\centering
	\includegraphics[width=3.3in]{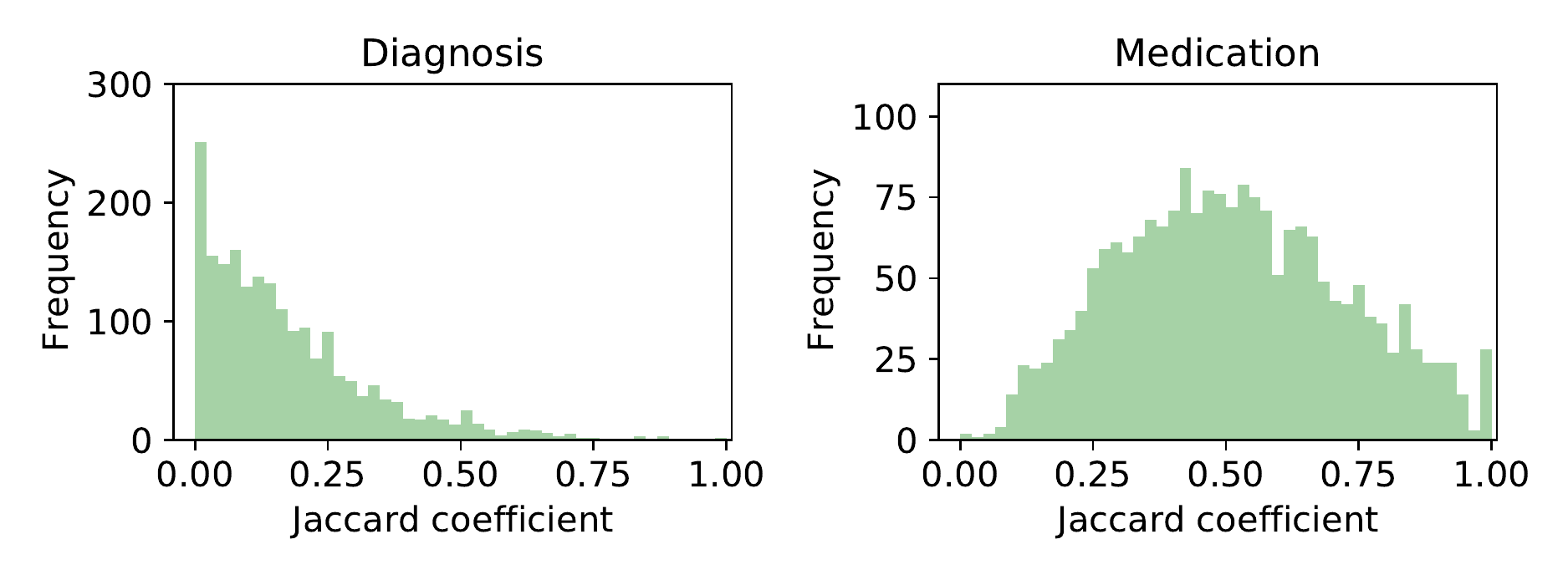}	\vspace{-3mm}
	\caption{Frequency Histogram of Jaccard Coefficients on {\em IQVIA}.}
	\vspace{-3mm}
	\label{fig:jaccard_distribution2}
\end{figure}

Here are the details about two datasets used in the paper.
\begin{itemize}
    \item (i) {\em MIMIC-III} \cite{johnson2016mimic} is a benchmark inpatient dataset, which collects electronic ICU records of Beth Israel Deaconess Medical Center between 2001 and 2012. We utilize the diagnosis, procedure, medication data and filter out patients with only one visit, so many patients will have 2 and more visits (up to 29 visits) in this data. From the original files in MIMIC-III ``DIAGNOSES\_ICD.csv", ``PROCEDURES\_ICD.csv" and ``PRESCRIPTIONS.csv", we extract diagnosis, procedure and medication code list, separately, then. These three files are then merged by the patient id and ``HADM\_ID", which is also used as visit id. After the merging, diagnosis and procedure are ICD-9 coded. 
    \item (ii) {\em IQVIA PharMetrics Plus}
% \footnote{iqvia.com/library/fact-sheets/iqvia-pharmetrics-plus} 
is a private outpatient dataset. Patients in this database is generally representative of the under 65 commercially insured population in the U.S., who receives treatment without being admitted to a hospital from 2015 to 2019. We keep visit records with at least $3$ medications. In this dataset, patients usually have around 10 visits. Similar to Figure~1 in the main text, we provide a Jaccard coefficient distribution for {\em IQVIA} dataset in Figure~\ref{fig:jaccard_distribution2}. Again, we observe that medications have larger overlaps than diagnosis.
\end{itemize}

The experimental settings are largely adopted from GAMENet \cite{shang2019gamenet} with minor adjustments. For both of the datasets, we extract the DDI information from Top-40 severity types from TWOSIDES \cite{tatonetti2012data}. We use ATC-3 Level drug coding to process all medications, use ICD-9 codes for diagnosis and procedures in {\em MIMIC-III}, and for {\em IQVIA}, we use ICD-10-CM codes for diagnoses and CPT/HCPCS codes for procedures. In MIMIC, ``visit" is identified by the field ``HADM\_ID" in raw data file. When running the experiments, we use multi-hot encodings for the diagnosis and procedure vectors.

%  Our models are not trained on data from one expert. They are trained on two large datasets of diagnosis and medications from many experts. Similar settings have been used in previous works,
 
%  Thanks for the suggestion. This is a well-known limitation of all existing works on medication recommendation. Even in practice, well-trained clinicians may not accurately predict medication effects on a given patient. As a practical compromise we adopt the same evaluation metrics as previous works.

% Our experimental 

%  Efficacy estimation requires labeled data which are difficult to acquire. We focus on medication prediction like many previous works. If such datasets become available, our model will still be applicable to evaluate the medication efficacy.

Two datasets are randomly split into training, validation and test by $60\% : 20\% : 20\%$. 
% \js{strange splits, by convention, we may want to have a split proportion in percentages but if this is indeed 4:1:1, then we can leave as it is }
% For the embeddings, GAMENet \cite{shang2019gamenet} uses 64-dim for their GRUs and other baselines do not explicitly mention the size. For fair comparison, we therefore implement all embedding sizes as 64-dim in the baselines. 
For {\em MIMIC-III}, we set embedding size $s=64$ for diagnosis and procedure tables, $\mathbf{E}_d,\mathbf{E}_p$. The health representation network, $\mbox{NET}_{health}$, is an affine transformation, we implement it as one linear layer without activation function. The prescription network, $\mbox{NET}_{med}$, is a two-layer feed-forward neural network with $256$ hidden units. The hyperparameters, $\gamma=0.75,\eta=0.08$, are selected from validation set. We choose RMSprop as the optimizer with learning rate at $2e^{-4}$ and weight decay at $1e^{-5}$.
For {\em IQVIA}, we set $s=32$, choose a three-layer $\mbox{NET}_{med}$ with $128$ hidden units, the hyperparameters $\gamma=0.75,\eta=0.03$, while the learning rate is $5e^{-4}$. For two datasets, the hyperparameters for the loss function are $\lambda_i=0.25,~i=1,2,3,4$. 

The implementation for baseline methods can be referred to RETAIN\footnote{https://github.com/mp2893/retain}, LEAP and GAMENet\footnote{https://github.com/sjy1203/GAMENet}. For MIMIC-III dataset, the hyperparameters of the baselines are mainly selected by grid search. 
\begin{itemize}
    \item For RETAIN, We choose two 64-dim GRUs as the implementation of two-level RNN with dropout rate at 0.5 for the output embedding. RMSprop is used as the optimizer with learning rate at $5\times 10^{-4}$ for RETAIN and other deep learning baselines. 
    \item LEAP is also implemented by 64-dim GRU as the feature encoder. However, we choose droput rate at 0.3 between two layers, as 0.3 works better than 0.5 in the validation set. Particularly, since LEAP are sequence-based models, we set 20 as the max drug combination size. 
    \item For GAMENet, we use the same suit of hyperparameters reported from the original paper \cite{shang2019gamenet}: expected DDI rate at 0.05, initial annealing temperature at 0.85, mixture weights $\pi=[0.9, 0.1]$ and also use 64-dim embedding tables and 64-dim GRU as RNN. We find the hyperparameter works well in validation set, so we keep it for test. 
\end{itemize} For {\em IQVIA}, the embedding size and hidden layers are reduced to 32-dim.
% \js{We can add more details and a diagram of the neural network in the supplement}
In both of the datasets, we could only observe a window of patient's visits, instead of a complete visit history. Since the evaluation is on medication change prediction, we assume the earliest visit appeared in the window as the ``first" visit of a patient $j$, and the evaluation of all models starts from the ``second" visit. All experiments are implemented by {\em PyTorch 1.4.0} on a Ubuntu server with 64GB memory, 32 CPUs and two GTX 2080Ti GPU. The reported results are averaged over 5 independent experiments with different random seeds.

\section{Details about Metrics}
This section provides the definition of each metric used in the experiment section.

We already present the calculation of {\em Err(add)} in the main text, in this section, we show the calculation of {\em Err(remove)}, DDI, Jaccard coefficient (Jaccard), F1-score (F1). All the metrics are calculated from the observed ``second" visits.

\paragraph{Err(remove).} The {\em Err(remove)} metrics for a particular patient $j$ is calculated by 
{\small\begin{align}
    \small \mathcal{O}_{target, j}^{(t)} &= \tilde{\mathcal{M}}_j^{(t-1)} \setminus \mathcal{M}^{(t)}_j\notag\\
    \mbox{Err(remove)}_j &= \frac{1}{V(j)}\sum_{t=2}^{V(j)} |\mathcal{O}_{target, j}^{(t)}\setminus \mathcal{O}_j^{(t)}| + |\mathcal{O}_j^{(t)}\setminus \mathcal{O}_{target, j}^{(t)}|\notag
\end{align}}
where $\mathcal{O}_{target}^{(t)}$ is the target removal set, $\mathcal{N}^{(t)}$ is the predicted removal set, $\mathcal{A}\setminus\mathcal{B}$ is set minus operation, $V(j)$ means total number of visits of patient $j$, $|\cdot|$ is for set cardinality, $|\mathcal{O}_{target,j}^{(t)}\setminus \mathcal{O}_j^{(t)}|$ and $|\mathcal{O}_j^{(t)}\setminus \mathcal{O}_{target,j}^{(t)}|$ are false negative and false positive numbers in terms of removal for the $t_{th}$ visit. The final {\em Err(remove)} is calculated by taking the average of all patients.

\paragraph{DDI rate.}
This metric is used to reflect the safety of drug combinations. 
As is stated in the main text, $\mathcal{M}_j^{(t)}$ is the target medication set and $\tilde{\mathcal{M}}_j^{(t)}$ is the estimated medication set for patient $j$ at the $t_{th}$ visit. The DDI rate for patient $j$ is calculated by,
\begin{equation} 
DDI_j = \frac{\sum^{V(j)}_{t=2}\sum_{m, n \in \tilde{\mathcal{M}}_j^{(t)}} \mathbf{1}\{\mathbf{A}_{mn}=1\}}{\sum^{V(j)}_{t=2}\sum_{m,n\in \tilde{\mathcal{M}}_j^{(t)}}1}, \notag
\end{equation} 
where $V(j)$ is the total number of visits of the $j$-th patient, $\mathbf{A}$ is the DDI matrix and $\mathbf{1}\{\cdots\}$ is an indicator function, which returns 1 when the expression in $\{\cdots\}$ is true, otherwise 0. The reported DDI rate is by taking average of all patients.

\paragraph{Jaccard.} 
The Jaccard coefficient for patient $j$ is calculated by 
\begin{equation}
    Jaccard_j = \frac{1}{V(j)}\sum_{t=2}^{V(j)}\frac{|\tilde{\mathcal{M}}_j^{(t)}\cap {\mathcal{M}}_j^{(t)}|}{|\tilde{\mathcal{M}}_j^{(t)}\cup {\mathcal{M}}_j^{(t)}|}, \notag
\end{equation}
while the overall Jaccard coefficient of the test data is by taking further average over all patients.
In the main text, we report only the overall Jaccard coefficient.

\paragraph{F1.} The F1 score is calculated by the harmonic mean of precision and recall. For patient $j$ at the $t$-th visit, the Precision, Recall, F1 are calculated by
\begin{equation}
    Precision_j^{(t)} = \frac{|\tilde{\mathcal{M}}_j^{(t)}\cap {\mathcal{M}}_j^{(t)}|}{ |\tilde{\mathcal{M}}_j^{(t)}|} \notag
\end{equation}
\begin{equation}
    Recall_j^{(t)} = \frac{|\tilde{\mathcal{M}}_j^{(t)}\cap {\mathcal{M}}_j^{(t)}|}{ |{\mathcal{M}}_j^{(t)}|} \notag
\end{equation}
\begin{equation}
    F1_j^{(t)} = \frac{2}{\frac{1}{Precision_j^{(t)}}+\frac{1}{Recall_j^{(t)}}} \notag
\end{equation}
The F1 score for one patient $j$ is by taking the average over his visits,
\begin{equation}
    F1_j = \frac{1}{V(j)}\sum_{t=2}^{V(j)}F1_j^{(t)}, \notag
\end{equation}
while the overall F1-score of the test data is by taking further average over all patients. In the main text, we report only the overall F1-score.

\section{Standard Deviation and $p$-value}
We provide the standard deviation and $p$-value results for performance comparison in MIMIC-III (Table~\ref{eq:mimic-iii-result}) and IQVIA (Table~\ref{eq:ds2-result}).
\begin{table*}[!htbp]
    \caption{Performance Comparison on {\em MIMIC-III}}
    \vspace{-2mm}
    \resizebox{\textwidth}{!}{\centering\begin{tabular}{c|ccccccc}
    \toprule
    \centering
      Metrics & DDI &Jaccard  & F1 & Err(add) & Err(remove)\\
         \midrule
        %  Repeat & 0.0816 / 0.3062 & 0.4739 / 0.2872 & 0.6327 / 0.3676 & 7.301 / 1.626 & 6.875 / 0.783 & --- & --- \\

SimNN & 0.0837 $\pm$ 0.0005 (2.5518e-11) & 0.4658 $\pm$ 0.0004 (6.6613e-16) & 0.6235 $\pm$ 0.0004 (4.4408e-16) & 6.856 $\pm$ 0.0054 (9.3702e-14) & 7.329 $\pm$ 0.0123 (1.5543e-15) \\

DualNN & 0.0925 $\pm$ 0.0006 (6.9011e-13) & 0.4880 $\pm$ 0.0004 (8.0380e-14) & 0.6447 $\pm$ 0.0004 (5.5733e-14) & 6.261 $\pm$ 0.0119 (7.6331e-07) & 7.987 $\pm$ 0.0112 (0.0) \\

RETAIN & 0.0932 $\pm$ 0.0016 (1.8352e-09) & 0.4796 $\pm$ 0.0006 (2.9309e-14) & 0.6389 $\pm$ 0.0005 (2.1316e-14) & 8.552 $\pm$ 0.0396 (2.4424e-15) & 6.338 $\pm$ 0.0369 (1.8950e-08) \\

LEAP & 0.0880 $\pm$ 0.0028 (3.2323e-06) & 0.4331 $\pm$ 0.0011 (4.4408e-16) & 0.5953 $\pm$ 0.0010 (4.4408e-16) & 9.105 $\pm$ 0.0356 (2.2204e-16) & 5.939 $\pm$ 0.0335 (0.0124) \\

GAMENet & 0.0928 $\pm$ 0.0007 (1.8409e-12) & 0.4980 $\pm$ 0.0012 (2.4894e-10) & 0.6549 $\pm$ 0.0011 (2.5575e-10) & 8.810 $\pm$ 0.1021 (4.8889e-12) & 5.854 $\pm$ 0.0450 (0.9780) \\
\midrule
\method & {0.0695} $\pm$ 0.0004  & {0.5234 } $\pm$ 0.0008& {0.6778 } $\pm$ 0.0007& {6.090} $\pm$ 0.0189& {5.853} $\pm$ 0.0219\\
    \bottomrule
    \end{tabular}} \label{eq:mimic-iii-result}
* $p$-values are at the parenthesis.
\end{table*}

\begin{table*}[!htbp]
    \caption{Performance Comparison on {\em IQVIA}}
    \vspace{-2mm}
    \resizebox{\textwidth}{!}{\centering\begin{tabular}{c|ccccccc}
    \toprule
    \centering
      Metrics & DDI &Jaccard  & F1 & Err(add) & Err(remove)\\
         \midrule
        %  Repeat & 0.0816 / 0.3062 & 0.4739 / 0.2872 & 0.6327 / 0.3676 & 7.301 / 1.626 & 6.875 / 0.783 & --- & --- \\
       SimNN & 0.0152 $\pm$ 0.0002 (0.0054) & 0.192 $\pm$ 0.0003 (0.0) & 0.2383 $\pm$ 0.0005 (0.0) & 2.906 $\pm$ 0.0031 (0.0) & 1.679 $\pm$ 0.0056 (1.0043e-12) \\

DualNN & 0.0156 $\pm$ 0.0003 (0.0015) & 0.112 $\pm$ 0.0003 (0.0) & 0.1783 $\pm$ 0.0003 (0.0) & 3.406 $\pm$ 0.0069 (0.0) & 2.579 $\pm$ 0.0051 (0.0) \\

RETAIN & 0.0279 $\pm$ 0.0008 (6.7339e-10) & 0.332 $\pm$ 0.0002 (1.5543e-15) & 0.4215 $\pm$ 0.0003 (4.4408e-16) & 2.353 $\pm$ 0.0229 (3.7266e-08) & 0.927 $\pm$ 0.0169 (better) \\

LEAP & 0.0134 $\pm$ 0.0013 (better) & 0.1871 $\pm$ 0.0007 (0.0) & 0.2742 $\pm$ 0.0008 (0.0) & 3.663 $\pm$ 0.0205 (2.2204e-16) & 1.286 $\pm$ 0.0153 (0.0005) \\

GAMENet & 0.0166 $\pm$ 0.0003 (1.7060e-05) & 0.2025 $\pm$ 0.0008 (0.0) & 0.3016 $\pm$ 0.0007 (0.0) & 3.016 $\pm$ 0.0589 (8.6569e-10) & 2.179 $\pm$ 0.0205 (9.7255e-14) \\
         \midrule
         \method &  0.0143 $\pm$ 0.0003 & {0.3634} $\pm$ 0.0005& {0.4544} $\pm$ 0.0004 & {2.088} $\pm$ 0.0104& 1.213 $\pm$ 0.0141\\
         \bottomrule
    \end{tabular}} \label{eq:ds2-result}
* $p$-values are at the parenthesis. \\
* In this dataset, for Err(remove) metric, RETAIN is better than our \method. For DDI, LEAP is better than our \method.
\end{table*}

\end{document}